\documentclass[twoside,11pt]{article}
\usepackage[preprint]{jmlr2e}
\usepackage{graphicx}
\usepackage{todonotes}
\usepackage[newfloat,frozencache,cachedir=minted-cache]{minted}
\usepackage{caption}

\jmlrheading{1}{2023}{1-48}{4/00}{10/00}{TurboGP}{Rodriguez-Coayahuitl, Morales-Reyes \& Escalante, HJ}

\ShortHeadings{TurboGP: A modern GP implementation}{Rodriguez-Coayahuitl, et al.}
\firstpageno{1}

\newenvironment{code}{\captionsetup{type=listing}}{}
\SetupFloatingEnvironment{listing}{name=Code}

\begin{document}

\title{TurboGP: A flexible and advanced python based GP library}
% combinacion de: TurboGP: A flexible and highly operational python based GP library y TurboGP: an advanced library for GP

\author{\name Lino Rodriguez-Coayahuitl \email linobi@inaoep.mx \\
        \name Alicia Morales-Reyes \email a.morales@inaoep.mx \\
        \name Hugo Jair Escalante \email hugojair@inaoep.mx \\
       %\addr Computer Science Department \\
       %Instituto Nacional de Astrof\'isica \'Optica y %Electr\'onica\\
       %Sta María Tonanzintla, Puebla, 72840, Mexico}
       \addr  Instituto Nacional de Astrof\'isica \'Optica y Electr\'onica\\
       Sta María Tonanzintla, Puebla, 72840, Mexico}

\editor{Kevin Murphy and Bernhard Sch{\"o}lkopf}

\maketitle

\begin{abstract}%   <- trailing '%' for backward compatibility of .sty file

We introduce \textit{TurboGP}, a Genetic Programming (GP) library fully written in Python and specifically designed for machine learning tasks. TurboGP implements modern features %in the GP framework, 
not available in other GP implementations, such as island and cellular population schemes, different types of genetic operations (migration, protected crossovers), online learning, among other features. TurboGP's most distinctive characteristic is its native support for different types of GP nodes %thus allowing GP implementations at different abstraction levels; 
to allow different abstraction levels, this makes TurboGP particularly useful for processing a wide variety of data sources. 
%\todo[author=HJ,inline]{Habr\'a que vincular explicitamente TurboGP o GP con machine learning, para que quede claro que estamos dentro del scope de JMLR}

\end{abstract}

\begin{keywords}
  Genetic Programming, Symbolic Regression,  Evolutionary machine learning, On-line learning.
\end{keywords}

\section{Introduction}

Genetic Programming (GP) is an evolutionary computation (EC) framework to automatically generate models and (simple) computer programs~\citep{koza1992genetic}. GP has been widely used for a variety of tasks including machine learning (ML), see e.g., \citep{guo2006breast,shao2013feature,cano2019evolving}. In GP, models are commonly represented by abstract syntax trees, such as the one depicted in Fig.~\ref{fig:GPTree}. GP's representation flexibility makes it appropriate to codify most models considered in machine learning problems, from classifiers~\citep{espejo2009survey} to reinforcement learning agents~\citep{co-reyes2021evolving}. GP works by initializing a population of randomly generated models, called individuals, then selects and transforms some of the best individuals in the population to generate a next generation of individuals that are better at solving the problem at hand. This process of selection $\rightarrow$ mutation $\rightarrow$ survival of the fittest, is repeated in an iterative cycle that mimics natural evolution. %This random search is guided by an \textit{objective function}, that assigns a \textit{fitness} value to each individual, i.e. to evaluate how good is every individual at the problem a user approaches through GP.

This paper introduces TurboGP, a GP library based on Python to target machine learning modeling problems. TurboGP implements standard components and techniques commonly used in GP, as well as recent developments in the field. 
%supports many proven techniques and upgrades that have been proposed through the years since the inception of GP in the research literature. 
We emphasize that many of these features are not found, or are difficult to re-implement, in available GP libraries (see Sec.~\ref{comparison}). Thus, we consider TurboGP a \textit{modern} GP implementation. The library is available at \href{https://github.com/l1n0b1/TurboGP}{https://github.com/l1n0b1/TurboGP}, where the code, notebook tutorials, as well as links to sample datasets needed for running the demos can be found.

\hspace{-.5cm}\begin{minipage}{\dimexpr\textwidth}

  \begin{minipage}[b]{0.49\textwidth}
    \centering
    \includegraphics[width=.50\textwidth]{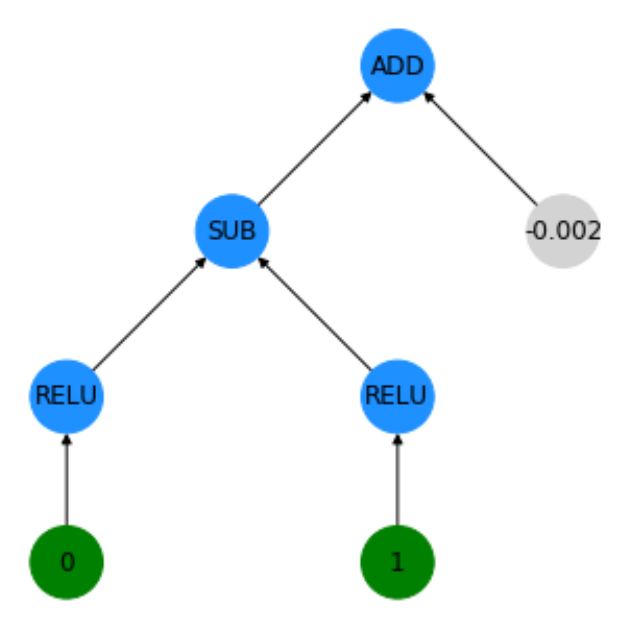}
    \captionof{figure}[GP individual tree]{\footnotesize{GP tree representing function $f(x_0,x_1)=(max(x_0,0)-max(x_1,0)) + (-.002)$}}
    \label{fig:GPTree}
  \end{minipage}
  \hfill
  \begin{minipage}[b]{0.49\textwidth}
    \centering
    \tiny{
\begin{tabular}{|l|c|c|c|}
\hline
   \textbf{Feature}                                 & \textbf{DEAP} & \textbf{gplearn} & \textbf{TurboGP} \\ \hline
\begin{tabular}[c]{@{}l@{}}Parallel \\processing  \end{tabular}                  & X             & X                & X                \\ \hline
\begin{tabular}[c]{@{}l@{}}Protected genetic \\operations   \end{tabular}        & X             & implicit         & X                \\ \hline
\begin{tabular}[c]{@{}l@{}}Different primitives \\layers \end{tabular}        & partial       & -                & X                \\ \hline
Online Learning                     & -             & partial          & X                \\ \hline
\begin{tabular}[c]{@{}l@{}}Spatially Distributed \\populations \end{tabular}     & -             & -                & X                \\ \hline
\end{tabular}}
      \label{Table:differences}
      \captionof{table}{\footnotesize{Features provided by different GP suites.}}
      
    \end{minipage}
\end{minipage}

%\begin{figure}
%\centering
%\begin{minipage}[t]{\dimexpr.5\textwidth-1em}
%  \centering
%    \includegraphics[width=.50\textwidth]{./figures/GPtreeLow}
%  \caption[GP individual tree]{\footnotesize{GP tree representing function $f(x_0,x_1)=(max(x_0,0)-max(x_1,0)) + (-.002)$}}
%  \label{fig:GPTree1}
%\end{minipage}\hfill
%\begin{minipage}[t]{\dimexpr.5\textwidth-1em}
%  \centering
%    \includegraphics[width=.70\textwidth]{./figures/algorithm}
%  \caption[GP individual tree]{\footnotesize{General procedure to implement a GP with TurboGP.}}
%  \label{fig:usage}
%\end{minipage}
%\end{figure}

\section{TurboGP}

TurboGP is designed in a modular fashion: core modules define classes and methods for basic building blocks, such as tree objects and fundamental genetic operations (e.g. subtree crossover, mutation); another core module defines the population dynamics, i.e. methods to perform evolutionary cycles under different survival/replacement policies, such as steady state; it also contains methods to implement cellular populations, where individuals are assigned a \textit{spatial location} property, to interact only (generate offspring, compete for survival) with individuals located within their neighborhood. GP individual classes define objects such as regression or classification models, i.e. the type of GP individuals to be evolved. These objects contain one or more GP trees, methods to evaluate them against a dataset, and variables to store their associated fitness value. GP individual classes also define wrappers for genetic operations that rely on methods implemented in core modules, but may also provide additional logic to implement more complex GP operations, such as crossovers that ensure offspring generated do not exceed max tree size limit, or that they are semantically valid (i.e., \textit{protected} crossovers). %Other modules provide selection mechanisms (\textit{tournament} and \textit{roulette-wheel} selection, \textit{elitism}) as well as "migration" operations, aimed at deploying multi-population (\textit{island} based) GPs~\citep{martin1997c6}. 
TurboGP also ships with a module that encapsulates and abstracts all its internal workings into a Scikit-alike interface, for rapid prototyping and experimentation. This modularity allows to easily modify and extend its functionality.

\subsection{Usage}
\label{sec:usage}

Code~\ref{code:example1} illustrates how TurboGP can be used to launch GP processes while combining standard GP steps and a Scikit-learn alike workflow: (lns.6-7) primitive sets declaration; (ln.8) GP-individual parameters setup, (lns.14-22) GP process instantiating and parameters declaration, i.e., individual class to evolve, genetic operations probabilities, pop size, selection mechanisms, etc.; (ln.28) GP evolutionary/learning run launch for a given training dataset.

\begin{code}
\begin{minted}[mathescape,
               fontsize=\footnotesize,
               linenos,
               numbersep=4pt,
               gobble=2,
               frame=lines,
               framesep=2mm]{python}             
  from genetic_program import GeneticProgram      # Module with scikit-alike interface
  from Regressor import RegressorLS               # GP individual we will use (Regressor)
    # ... (dataset preparation)
  lowlevel = ['ADD', 'SUB', 'MUL', 'DIV', 'RELU', 'MAX', 'MEAN', 'MIN', 'X2', 'SQRT'] 
  GeneticProgram.set_primitives(lowlevel=lowlevel) # Primitives
  ind_params = {'input_vector_size':2, 'complexity':12} # parameters for individuals to evolve   
  # Genetic operations to use.
  oper = [RegressorLS.mutation, RegressorLS.protected_crossover, RegressorLS.mutation_i2]      
  oper_prob = [.4, .4, .2]                        # operations probabilities
  oper_arity = [1, 2, 1]                          # operations arity (parents required)
  GP = GeneticProgram(individual_class=RegressorLS,       # individual type to evolve
                      operations=oper, operations_prob=oper_prob,
                      operations_arity=oper_arity, ind_params=ind_params,                                  
                      pop_size=1000,                      # population size
                      generations=100,                    # no. generations to elapse
                      pop_dynamics="Steady_State",        # population dynamic to use
                      online=False,                       # if online learning mode enabled
                      minimization=True,                  # minimization or maximization problem
                      n_jobs=16)                          # CPU/threads to use 
    gp_reg = GP.fit(x_training, y_training)                 # launch training/evolution
\end{minted}
\vspace*{-\baselineskip}
\caption{\footnotesize{TurboGP example for evolving a regression model through the Scikit-learn alike interface.}}
\label{code:example1}
\end{code}

\normalsize

\subsection{Main features}

TurboGP supports the following characteristics at processing and algorithmic levels:

\begin{itemize}
\item \textbf{Parallel processing.} TurboGP can use multiple CPU cores/threads in different ways: parallel individuals evaluation, and multiple populations evaluation.

%\item \textbf{Protected max depth genetic operations.} TurboGP provides two protected subtree crossover variants: \texttt{protected\_crossover} and \texttt{samedepths\_crossover}~\citep{harries1997exploring}. In addition, \texttt{subtree\_mutation} is protected by default. %These protected genetic operation variants help counteract \textit{bloat}.

\item \textbf{Different primitives layers.} TurboGP allows trees representation that may contain different types of nodes/primitives, e.g. scalar (\textit{low} level) functions, vector-to-scalar (\textit{mezzanine}) functions, vector-to-vector functions, etc. These kind of models are useful in high dimensional learning problems~\citep{al2012two,evans2018evolutionary}.%,rodriguez2019comparison}.

\item \textbf{Explicit support for on-line learning} (also called incremental or mini-batch learning). TurboGP explicitly emphasises whether on-line learning mode is turned on, to increase efficiency and improve convergence~\citep{rodriguez2019evolving}. 

\item \textbf{Spatially distributed populations}. TurboGP supports models to allocate individuals in toroidal grid arrangements (cellular)~\citep{petty1997diffusion}, as well as migration operations to implement multi-population (island) models~\citep{martin1997c6}.
\end{itemize}

\subsection{Related available libraries}
%\section{Comparison to other software}
\label{comparison}
Several available GP libraries cannot be compared to TurboGP due to critical differences, such as not being written in Python (ECJ, \cite{scott2019ecj}) or require propietary software to run (GPLAB,~\cite{silva2003gplab}). Among those available as Python libraries, some are too minimalistic (TinyGP,~\cite{Sipper2019tinyGP}), not modular nor designed to be modified (Karoo GP,~\cite{staats2017tensorflow}) or are implemented in older versions of Python (pySTEP, ~\cite{khoury2010using}). DEAP~\citep{fortin2012deap} and gplearn~\citep{stephens2019gplearn} are two recent Python GP libraries that are comparable to TurboGP. %DEAP is a library aimed at EC in general, that supports GP, while gplearn is GP specific. 
DEAP, gplearn and TurboGP, support basic features, such as easy primitives declaration and graphic individuals visualization. However, TurboGP provides new features and strengthen others partially supported by those libraries. %For example, while both DEAP and TurboGP support max depth protected genetic operations, gplearn only provides a functionally similar behavior through a parsimony coefficient that penalizes large individuals' fitness. Similarly, while gplearn allows to evaluate individuals with subsets of the training set, TurboGP goes further, as it provides an explicit on-line learning mode.
Table~\ref{Table:differences} summarizes the differences.

%DEAP~\citep{fortin2012deap} and gplearn~\citep{stephens2019gplearn} are the two highest ranked, and recent Python GP libraries available on GitHub. DEAP is a library aimed at EC in general, that supports GP, while gplearn is GP specific. All three, DEAP, gplearn and TurboGP, support basic features, such as easy primitives declaration and graphic individuals visualization. However, TurboGP provides new features and strengthen others partially supported by those libraries. For example, while both DEAP and TurboGP support max depth protected genetic operations, gplearn only provides a functionally similar behavior through a parsimony coefficient that penalizes large individuals' fitness. Similarly, while gplearn allows to evaluate individuals with subsets of the training set, TurboGP goes further, as it provides an explicit on-line learning mode. Table~\ref{Table:differences} summarizes the differences.

%DEAP allows declaring strongly typed GP individuals, a feature that could be used to declare different primitives classes; TurboGP goes further, as it provides primitives at different abstraction levels (runtime variable size input/ouput primitives). Table~\ref{Table:differences} summarizes the differences. %\textcolor{red}{Finally, the features that are unique to TurboGP and have proven to be quite useful for solving machine learning problems are: online learning support, X, Y, Z.}

\begin{table}[]
\caption{\footnotesize{Parameters used for different tasks and GP setups tested with TurboGP.}}
\label{Table:setup}
\centering
\scriptsize{
\begin{tabular}{|l|cccccc|}
\hline
                              & \multicolumn{2}{c|}{\textbf{Classification}}                                 & \multicolumn{2}{c|}{\textbf{Regression}}                                          & \multicolumn{2}{c|}{\textbf{Denoising}}                \\ \hline
                              & \multicolumn{1}{c|}{\textit{Batched}} & \multicolumn{1}{c|}{\textit{Online}} & \multicolumn{1}{c|}{\textit{Panmictic}} & \multicolumn{1}{c|}{\textit{Multi-Pop}} & \multicolumn{1}{c|}{\textit{Low}} & \textit{Mezzanine} \\ \hline
\textbf{Total Pop size}       & \multicolumn{2}{c|}{500}                                                     & \multicolumn{2}{c|}{4000}                                                         & \multicolumn{2}{c|}{1000}                              \\ \hline
\textbf{\# Populations}       & \multicolumn{2}{c|}{1}                                                       & \multicolumn{1}{c|}{1}                  & \multicolumn{1}{c|}{16}                 & \multicolumn{2}{c|}{1}                                 \\ \hline
\textbf{Dataset(Batch) size} & \multicolumn{1}{c|}{1200(1200)}      & \multicolumn{1}{c|}{1200(60)}       & \multicolumn{2}{c|}{5000(100)}                                                   & \multicolumn{2}{c|}{12000(200)}                       \\ \hline
\textbf{Generations}          & \multicolumn{1}{c|}{20}               & \multicolumn{1}{c|}{40}              & \multicolumn{2}{c|}{100}                                                          & \multicolumn{2}{c|}{60}                                \\ \hline
\textbf{Max tree depth}       & \multicolumn{2}{c|}{6}                                                       & \multicolumn{2}{c|}{12}                                                           & \multicolumn{2}{c|}{9}                                 \\ \hline
\textbf{Genetic Operations}   & \multicolumn{6}{c|}{subtree mutation, subtree crossover (protected), numeric mutation}                                                                                                                                                \\ \hline
\textbf{Operations Probs}     & \multicolumn{2}{c|}{(.5, .5, .0)}                                            & \multicolumn{2}{c|}{(.4, .4, .2)}                                                 & \multicolumn{2}{c|}{(.5, .5, .0)}                      \\ \hline
\textbf{CPU Threads}          & \multicolumn{2}{c|}{2}                                                       & \multicolumn{2}{c|}{16}                                                           & \multicolumn{2}{c|}{8}                                 \\ \hline
\textbf{Primitives (scalar)}  & \multicolumn{6}{c|}{$+$, $-$, $\times$, $\div$, max, min, mean, ReLU, $a^2$, $\sqrt{a}$}                                                                                                                                  \\ \hline
\textbf{Primitives (vector)}  & \multicolumn{5}{c|}{N/A}                                                                                                                                                                             & mean, min, max  \\ \hline
\end{tabular}}
\end{table}

\section{Benchmarking}
\label{benchmarks}

\begin{figure}[h!]
    %\hspace{-1.5cm}\includegraphics[width=1.32\textwidth]
    \includegraphics[width=1\textwidth]{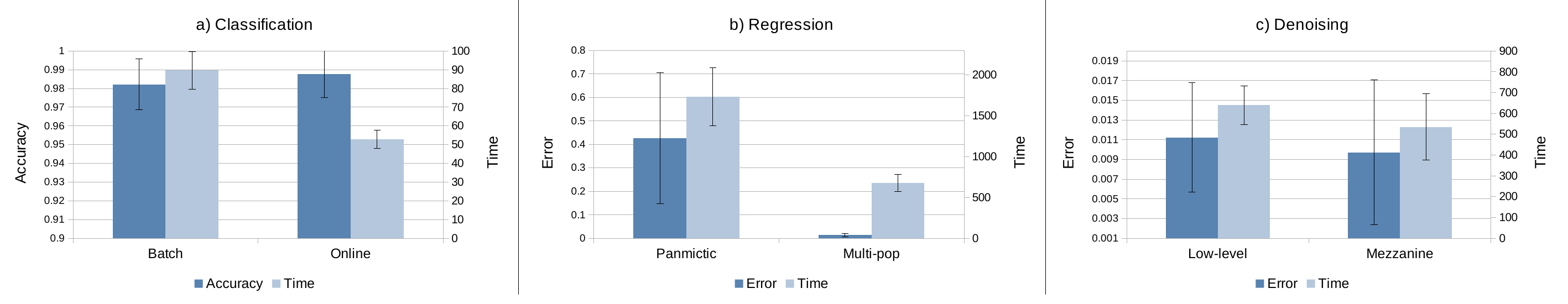}
  \caption[GP individual tree]{\footnotesize{Average results from 30 independent runs and different setups. For a) vertical axis is test accuracy (error for b and c), hence higher (lower) is better. Mean exec time expressed in seconds, lower is better for all cases. All experiments performed on a AMD Ryzen 7 1700 CPU on Debian GNU/Linux 11.}}
  \label{fig:results}
\end{figure}

The benefits of TurboGP features highlighted previously
%, and that are absent in similar software libraries. Specifically, we focus on demostrating
demonstrate the performance gains from on-line learning, multi-population models, and \textit{mezzanine} type of primitives. We selected three different ML tasks: classification, regression and image denoising. For classification we used the banknote authentication dataset~\citep{lohweg2012banknote}  from the UCI repository~\citep{Dua2019}, with  1372 samples where we used 1200 (172) for training (testing). For regression, we generated 5000 (500) training (testing) samples using "Keijzer 12" function~\citep{keijzer2003improving}, $f(z) = xy + \sin{((x-1)(y-1))}$. For the denoising task, 14,000 image patches of $21 \times 21$ pixels in size were extracted from BSDS~\citep{MartinFTM01}, and contaminated with additive noise; we used 12,000 (2000) patches as training (testing) set, and setup a GP to find a model capable of cleaning the image patches. Table~\ref{Table:setup} lists in detail the different parameters for each GP run. Fig.~\ref{fig:results} shows the results obtained by the different GP setups. Results show that online learning allows to decrease GP runtime without taking any toll on classification performance; multi population schemes drastically reduce both, convergence time, as well as regression error; and vector-to-scalar primitives allow GP to reach better solutions than scalar-only GP primitives.

\section{Forthcoming features}

The library is being constantly updated. The next major features we plan to incorporate into TurboGP are: automatically defined functions~\citep{koza1994genetic}, memetic GP representations~\citep{emigdio2014evaluating} and co-evolutionary algorithms. We already have a Cooperative Coevolutionary framework in the development branch~\citep{rodriguez2020cooperative}.

%\newpage
% Acknowledgements should go at the end, before appendices and references

\acks{This work was supported by project grant CONACYT CB-S-26314. Lino Rodriguez acknowledges support for this project
from Consejo Nacional de Ciencia y Tecnlog\'ia (CONACYT) grant No. 436184, Consejo de Ciencia y Tecnolog\'ia del Estado de Puebla (CONCYTEP) grant 2019-52D, and Instituo Nacional de Astrof\'isica, \'Optica y Electr\'onica "Beca de Colaboraci\'on 2020" grant. }

\bibliography{mybib}

\end{document}